\pdfoutput=1
\documentclass[11pt]{article}

\usepackage[preprint]{coling}

\usepackage{times}
\usepackage{float} 
\usepackage{placeins} 

\usepackage{latexsym}
\usepackage{booktabs}

\usepackage[T1]{fontenc}

\usepackage[utf8]{inputenc}

\usepackage{microtype}

\usepackage{inconsolata}

\usepackage{graphicx}

\usepackage{multirow}

\title{Lightweight Safety Guardrails Using Fine-tuned BERT Embeddings}

\author{ Aaron Zheng \\
  Uniphore \& UC Berkeley\\
  \texttt{aaron.zheng@uniphore.com} \\
  \texttt{aaronz@berkeley.edu}\\\And
   Mansi Rana\\
  Uniphore \\
    \texttt{mansi.rana} \\
      \texttt{@uniphore.com}\\\And
   Andreas Stolcke\\
  Uniphore \\
  \texttt{andreas.stolcke} \\
  \texttt{@uniphore.com}\\}

\begin{document}
\maketitle
\begin{abstract}
With the recent proliferation of large language models (LLMs), enterprises have been able to rapidly develop proof-of-concepts and prototypes. As a result, there is a growing need to implement robust guardrails that monitor, quantize and control an LLM's behavior, ensuring that the use is reliable, safe, accurate and also aligned with the users' expectations. Previous approaches for filtering out inappropriate user prompts or system outputs, such as LlamaGuard and OpenAI’s MOD API, have achieved significant success by fine-tuning existing LLMs. However, using fine-tuned LLMs as guardrails introduces increased latency and higher maintenance costs, which may not be practical or scalable for cost-efficient deployments.  We take a different approach, focusing on fine-tuning a lightweight architecture: Sentence-BERT. This method reduces the model size from LlamaGuard’s 7 billion parameters to approximately 67 million, while maintaining comparable performance on the AEGIS safety benchmark.

\end{abstract}

\section{Introduction}
The challenge of creating reliable, safe, accurate, and user-aligned automated knowledge retrieval systems has been a longstanding one, extensively studied since the advent of the internet. Much of the prior research has focused on search engines and enterprise search software \cite{salton1989,cutting1993,brin1997,broder2002}. Significant efforts have been made toward developing effective content moderation and filtering methods, leading to products, such as Google's SafeSearch, Bing SafeSearch, YouTube's Restricted Mode, Facebook's Community Standards filters, and Twitter’s Trust \& Safety tools. These solutions aim to provide users with safer and more curated content, while balancing accuracy and user expectations. 

Recently, there has been a surge of LLM-based guardrails, such as LlamaGuard, which aim to improve the safety, reliability, and control of LLMs in various applications. These guardrails work by utilizing fine-tuned LLMs either directly or as a classifier to filter out unsafe prompts, providing developers with more granular control over how LLMs interact with users in real-world applications. These advancements represent a growing effort across the AI industry to build responsible AI systems that are safer and more trustworthy.

However, using LLM-based guardrail models introduces high latency and significant inference costs, often requiring expensive GPU resources and substantial processing time. As compute costs remain high, the heavy-duty nature of current LLM guardrails can limit their use in cost-limited use cases. Such LLM application scenarios could include classrooms, private settings, or task automation in small businesses. The lack of effective and lightweight guardrail solutions for such uses represents a significant vulnerability, impacting both users and society as a whole.

We propose a lightweight guardrail solution by fine-tuning models based on BERT (bidirectional encoder representations from transformers) \cite{kenton2019bert} for effective unsafe prompt filtering.\footnote{The same method could be used to filter LLM outputs, but in this paper we focus on guarding against dangerous LLM inputs.}
The goal of our model is to classify whether a user's prompt or a conversation snippet is safe or unsafe. To achieve this, we fine-tune a BERT-based model on labeled safe and unsafe inputs, aiming to cluster safe and unsafe embedding vectors separately. We then train a classifier on these embedding vectors to discriminate between safe and unsafe content. Despite the simplicity of this approach, we demonstrate performance comparable or superior to that of more resource-intensive LLM-powered guardrail checkers.

Our approach frames the safety task purely as a text (e.g., topic) classification problem. For unsafe prompt filtering, we utilize the learned embedding model to convert each prompt into a vector representation within a high-dimensional space. In the paper, we discuss related work, describe our training procedure, present our results, and analyze areas of improvement and future steps.

Our contribution is twofold.  First, develop
an innovative, low-cost approach for building simple guardrails for LLMs, able to filter unsafe prompts effectively, thereby laying the groundwork for widespread use of safety models, and enabling developers to add those guardrails to their products.

Our second contribution is to show that our light-weight method has results comparable to state-of-the-art benchmarks in this space, such as LlamaGuard and OpenAI MOD API, on the AEGISSafetyDataset for safe versus unsafe prompt classification.
   
\section{Related Work}

\subsection{LlamaGuard}


LlamaGuard is a guardrails model published by Meta, and is an LLM-based solution for human-AI conversation use cases \cite{inan2024}. LlamaGuard is created by instruction-tuning Llama2-7b on an in-house safety dataset with ~14k training examples of possible human and AI assistant conversations. Data is labeled either as safe, or one of 6 risk categories: Criminal Planning, Suicide \& Self Harm, Regulated or Controlled Substances, Guns \& Illegal Weapons, Sexual Content, and Violence and Hate. LlamaGuard has 7 billion parameters, and achieves comparable performance to OpenAI API and Perspective API guardrails for the Toxic Chat and OpenAI Mod Datasets. According to the authors, LlamaGuard achieves superior performance on in-house safety datasets (but which they have not released publicly).

\subsection{NeMo}

NeMo 43B \cite{rebedea2023} is a model published by Nvidia, with 43 billion parameters, trained on 1.1 trillion tokens, spanning a diverse corpus of web-crawl, news articles, books, and scientific publications. This model is not developed primarily used for guardrails, but for language understanding tasks. However, Nvidia has produced NeMo43B-DEFENSIVE, a model fine-tuned on their open-sourced safety dataset, AEGIS \cite{ghosh2024} (see Section\ \ref{aegis} below).

\subsection{OpenAI MOD API}
OpenAI MOD API \cite{openaiAPI} is a closed-source API created by OpenAI for moderating the prompt and responses returned by LLMs. It can be used to classify both text and image-based inputs, and contains a diverse taxonomy of unsafe categories. The API itself can be used by developers by giving text as input, and the API will output a dictionary containing the probabilities of the input being in each unsafe category.

\subsection{Perspective API}
Perspective API \cite{lees2022} is an API developed by Jigsaw (part of Google) that detects toxic content in conversations. Like OpenAI, it is a closed-source model that can provide probability scores for unsafe categories, but it offers a public API that developers can use to analyze their conversations and highlight toxic content. 

\subsection{WildGuard}
WildGuard \cite{han2024} is a lightweight LLM-based safety model created through instruction tuning on the ``Mistral-7b-v0.3'' model. Its goal is to identify malicious intent in user prompts, assess safety risks of model responses, and determine model refusal rates. Like LlamaGuard, this model has 7 billion parameters. It required around 5 hours to train on four A100 80GB GPUs; it is one of the most recent safety models. 

\subsection{AEGIS fine-tuned models}
    \label{aegis}
AEGIS \cite{ghosh2024} is an AI content safety moderation solution developed by Nvidia, offering three distinct guardrail approaches along with an annotated, quality-assured safety dataset featuring a custom taxonomy. This taxonomy includes a special ambiguous category, "Needs Caution," which can be classified as either safe or unsafe. The three guardrail models are: (1) \textbf{LlamaGuard-Permissive}, which is instruction-tuned on their safety dataset, treating the ambiguous category as safe; (2) \textbf{LlamaGuard-Defensive}, similar to LlamaGuard-Permissive but treating the ambiguous category as unsafe; and (3) \textbf{NeMo-43B-Defensive}, derived by instruction-tuning on NeMo-43B and also treating the ambiguous category as unsafe. For evaluation purposes, the ambiguous samples are always treated as unsafe \cite{aegis-personal}.

\begin{figure*}[t]
    \centering
    \includegraphics[width=0.8\textwidth]{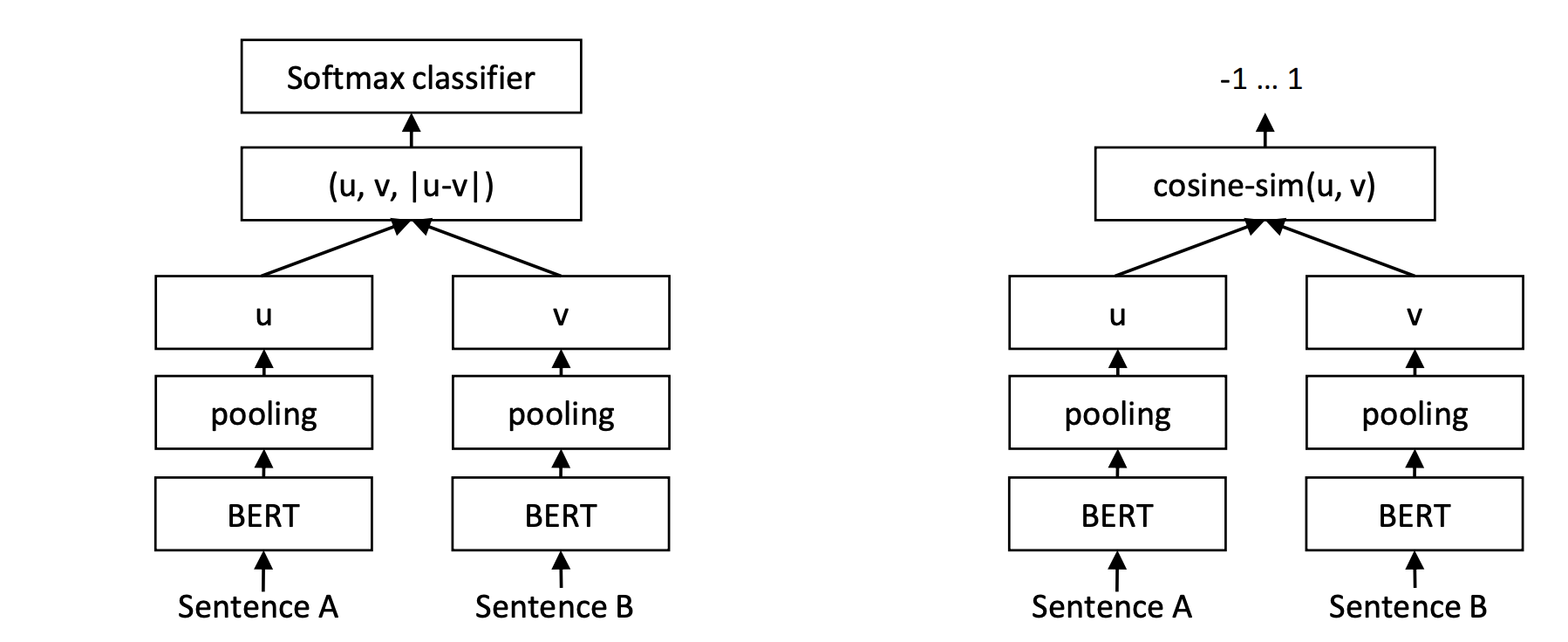} 
    \caption{Sentence transformer architecture.                  
    Left: training. Right: inference.}
    \label{fig:Sentence Transformer Architecture}
\end{figure*}

\section{Methodology}

\subsection{Problem Statement}
Given an input text $T$, which may contain individual user prompts or a conversation between a user and an agent, we want to be able to classify the input text as either \textit{safe} or \textit{unsafe}.

\subsection{Data}

To train our embedding model, we need a dataset containing text-based information labeled as safe or unsafe. Ideally, the genre of text should match the conversations between users and LLM.  Unfortunately, that there is a severe shortage of such well-labeled safety datasets.

The AEGISSafetyDataset is one such properly labeled dataset, an English-language corpus released by Nvidia containing ``approximately 26,000 human LLM interaction instances complete with human annotations'' \cite{ghosh2024}. The annotations use a taxonomy containing one broad safety category, and 13 critical risk areas. The risk areas are: Criminal Planning/Confessions, Identity Hate, Sexual, Violence, Suicide and Self-harm, Threat, Sexual (Minor), Guns/Illegal Weapons, Controlled/Regulated Substances, Privacy, Harassment, Needs Caution, and Other. We chose this dataset since we believe it to be the currently most comprehensively annotated publicly available safety dataset. 

\subsection{Embedding Model}

Sentence-BERT \cite{reimers2019} is a well-documented approach for generating embedding vectors for sentences, building upon the standard BERT model \cite{kenton2019bert}. It significantly enhances the efficiency in representing similarity and differences between texts using BERT-based models. The original BERT architecture lacks a mechanism to compute independent vector embeddings for sentence comparisons, resulting in substantial latency for sentence similarity tasks, as both sentences must be fed as a sequence into the BERT encoder to evaluate their similarity.

To solve this problem, Sentence-BERT uses a Siamese architecture. First, two texts are fed into two copies of the same BERT model. Then, vector embeddings from both BERT models are separately pooled into one embedding per sentence, resulting in two embeddings. Finally, the resulting embeddings are subjected to a loss function based on softmax (in training), or cosine similarity (for inference), as in Figure~\ref{fig:Sentence Transformer Architecture}. The loss is propagated equally through both Siamese BERT encoders, and the gradients are aggregated. With Sentence-BERT, the task of comparing sentence similarity is reduced to computing the cosine similarity of two embedding vectors, resulting in a substantial reduction of the computational overhead.

To obtain sentence-level embedding, three different methods are considered to obtain a single vector from BERT token embeddings. The simplest approach is to use the CLS token embedding, the embedding vector of the BERT special token that is used for next-sentence prediction. The two other methods consist of taking the maximum and the average, respectively, over all BERT token embeddings in the sentence.

While fine-tuning, the Sentence-BERT framework uses one of two loss functions: contrastive loss and triplet loss. Contrastive loss takes in pairs of sentences as input, labeled as either 1 (similar) or 0 (dissimilar), and the loss is computed as the difference between the softmax result and the labeled value. Triplet loss, on the other hand, takes in individual sentences labeled with integers ($0, 1, 2, \ldots, n$) representing the $n$ different categories. These individual sentences are separated into batches, used to form all possible sample triplets of (anchor, positive, negative), and each possible pair's embedding distance is either maximized (for positive, or same-label pairs) or minimized (for negative, or different-label pairs). Within triplet loss, there exists three variations that we explored: 1. BatchAll minimizes the combined sum of triplet losses per batch; 2. BatchHardMargin minimizes loss for the triplet with maximum loss per batch; and 3. BatchHardSoftMargin sums triplet losses for triplets, with each triplet's loss calculated as $max(0, d(A,P)-d(A,N)+margin)$, where $d$ is the distance, and $A, P, N$ the anchor, positive and negative embedding vectors, respectively. The default margin used in this framework is $1.0$.

\begin{table}[tb]
\centering
\resizebox{0.5\textwidth}{!}{
\begin{tabular}{|p{6cm}|c|}
\hline
\textbf{Category} & \textbf{Training data instance count} \\
\hline
Controlled/Regulated Substances & 417 \\
Criminal Planning/Confessions & 1824 \\
Fraud/Deception & 1 \\
Guns and Illegal Weapons & 179 \\
Harassment & 711 \\
Hate/Identity Hate & 848 \\
PII/Privacy & 510 \\
Profanity & 241 \\
Safe & 3217 \\
Sexual & 340 \\
Sexual (minor) & 27 \\
Suicide and Self Harm & 51 \\
Threat & 22 \\
Violence & 249 \\
\hline
\textbf{Total} & \textbf{8637} \\
\hline
\end{tabular}
}
\caption{Training data instance counts by category}
\label{Training Data}
\end{table}

\begin{table}[tb]
\centering
\resizebox{0.5\textwidth}{!}{
\begin{tabular}{|p{6cm}|c|}
\hline
\textbf{Category} & \textbf{Testing data instance count} \\
\hline
Controlled/Regulated Substances & 58 \\
Criminal Planning/Confessions & 232 \\
Fraud/Deception & 0 \\
Guns and Illegal Weapons & 22 \\
Harassment & 83 \\
Hate/Identity Hate & 95 \\
PII/Privacy & 47 \\
Profanity & 26 \\
Safe & 401 \\
Sexual & 34 \\
Sexual (minor) & 2 \\
Suicide and Self Harm & 7 \\
Threat & 4 \\
Violence & 26 \\
\hline
\textbf{Total} & \textbf{1037} \\
\hline
\end{tabular}
}

\caption{Test data instance counts by category}
\label{Testing Data}
\end{table}

\subsection{Overall Architecture}

We formulate the task of creating LLM guardrails as a two-stage architecture. Given a user input prompt text $T$, the first stage processes $T$ through an embedding model, fine-tuned on training data with corresponding labels (safe vs unsafe). The goal of this embedding model is to effectively learn to differentiate between safe inputs and unsafe inputs. In the second stage, a classifier takes the embedding vector output from the first stage and classifies it as either safe or unsafe. If the user prompt is safe, then we pass the user prompt as input to the LLM. If not, we output a generic response, telling the user that their input is unsafe, or otherwise refuse to engage with the prompt.

We utilize the "distilbert-base-uncased" model implementation from the Huggingface Transformers library as our BERT model. This model is smaller and faster than the original BERT, featuring 6 layers, each with a 768-dimensional hidden layer and 12 attention heads. Our choice of "distilbert-base-uncased" is motivated by three factors. First, it is a BERT-based model, pretrained on tasks such as next sentence prediction (NSP) and masked token prediction, giving it a strong grasp of word and sentence contexts through word and CLS token embeddings. Second, it is, to our knowledge, the smallest model that delivers comparable performance to other BERT-based models \cite{huggingface}.
Lastly, this model has not been fine-tuned on natural language inference tasks, which are not relevant, or could even be counter-productive, for our use case, in contrast to other models such as ``distilbert-base-uncased-finetuned-mnli'' and ``all-MiniLM-L6-v2'' \cite{sentence-transformers}.\footnote{For example, embeddings that support detection of logical contradictions are not helpful to our similarity learning since logically contradictory statements are likely to be in the {\em same} topical category.}

We then fine-tune a Sentence-BERT model, using the chosen underlying BERT model, for a predefined number of epochs and batch size to serve as our embedding model. After fine-tuning, we generate embeddings for both training and test data. For classification, we evaluate two model types: support vector machine (SVM) and a (shallow) neural network. Both models will be trained using the training data embeddings. Model accuracy will be assessed by running the best-performing classifier on the test embeddings and comparing to the corresponding labels.

\section{Data Preprocessing}
Each node in the AEGIS corpus is annotated at least three times by different annotators.
Sometimes, different annotators may disagree, either in whether data is safe or unsafe, or the specific unsafe taxonomies that are included. Since the dataset provides the individual annotator labels, we had to devise a label reconciliation scheme for classifier training and evaluation.

To get the final dataset labeled with AEGIS' custom taxonomy, we first take the second annotator's label without loss of generality. Then, we check for data labeled ``Other'', and see if another annotator labeled it as something other than ``Other''. If so, we replace the ``Other'' label. If not, we remove the data associated with the label. Then, for every ``Safe'' label, we want to be sure that the data is really safe. So, for ``Safe'' data, we check the other annotations to see if other annotators agree and label the data as ``Safe''. If so, we label the final data ``Safe''. If not, we label it as one of the unsafe categories that the other annotators annotated, randomly. For some data that may have more than one labels, we choose the first label, unless that label is ``Safe'', in which case we choose an unsafe label at random. Once all data is labeled with exactly one label, we remove all data with the ``Needs Caution'' label from both train and test, as it is ambiguous whether or not such data is safe or unsafe.  We also note that there are some duplicate instances (i.e., two or more data items with the same prompt string) in the publicly available AEGIS data, which we process by retaining only the first occurrence of said item. The ``first occurrence'' here follows the original indices in the AEGIS dataset.

It should be noted that most train and test inputs are within BERT's limit of 512 tokens. We use the default tokenization model of ``distilbert-base-uncased'' to tokenize inputs; this defaults to truncation of samples exceeding the limit.

The category distribution of training and test data after preprocessing is listed in Tables~\ref{Training Data} and~\ref{Testing Data}. The amount of data we use is somewhat less than that used by the AEGIS authors to instruction-tune their models. Our number of fine-tuning samples is 9,674, comparatively less than about 13,000 instances as used by AEGIS \cite{ghosh2024}.

\subsection{Classifier Setup}
    \label{classifier-setups}

To perform the final binary safe/unsafe classification, we experimented with four different setups:
\begin{itemize}
    \item Binary Embedding, Binary Classification (BEBC): Both the training and test datasets are divided into two categories, ``safe'' and ``unsafe'', with the latter consisting of all samples not labeled ``safe''. We fine-tune a single Sentence-BERT model and train a single binary classifier on these embeddings. 
    \item Multiple Embedding, Multiple Classifiers (MEMC): Treat each unsafe category as distinct. We then fine-tune seven Sentence-BERT models: ``safe'' against each category from a subset of populated unsafe categories (Criminal, Privacy, Sexual, Harassment, Guns, Violence, Control), and train a classifier for each fine-tuned model. The final label is deemed safe if all seven classifiers agree, and unsafe otherwise. 
    \item Multi-class Embedding, Multiple Classifiers (McEMC): Fine-tune a Sentence-BERT model, treating safe and each category of unsafe as distinct, and train seven classifiers, using the subsets from MEMC. Like MEMC, we aggregate results, making the final label safe only if all seven classifiers agree. 
    \item Multi-class Embedding, Multi-class Classification (McEMcC): This approach is similar to McEMC, where we fine-tune a Sentence-BERT model by treating ``safe'' and each unsafe category as distinct. However, instead of training seven separate classifiers, we use a single multi-class classifier to differentiate between all categories.
\end{itemize}

\section{Results}

To decide which classifier setup was most effective, we trained every embedding model for 10 epochs and implemented the four schemes with SVM classifiers (binary or multi-class, as needed).
See Appendix~\ref{subsec:hyperparameter_lis} for the full list of hyperparameters.
As shown in Table~\ref{Results Comparison},
``Multi-class Embedding, Multi-class Classification'' (McEMcC) outperformed the other classifier approaches.

\begin{table}[tb]
\centering
\resizebox{0.5\textwidth}{!}{
\begin{tabular}{|l|c|c|c|c|}
\hline
\textbf{Approach} & \textbf{BEBC} & \textbf{MEMC} & \textbf{McEMC} & \textbf{McEMcC} \\
\hline
\textbf{Accuracy} & 87.46\% & 84.67\% & 88.04\% &\bf 88.14\% \\
\hline
\textbf{F1 score} & 86.56\% & 84.91\% & 86.99\% &\bf 87.06\% \\
\hline
\textbf{UAP} & 86.69\% & 84.92\% & 87.16\% &\bf 87.24\% \\
\hline
\end{tabular}
}
\caption{Results comparison with different classifier setups. See Section-\ref{classifier-setups} for the meaning of the shorthands BEBC, BEMC, McEMC, McEMcC. UAP = unweighted average precision, a macro-averaged accuracy.}
\label{Results Comparison}
\end{table}

Once we had settled on the McEMcC classifier approach, we tested a variety of hyperparameters, including the choice between SVM and neural network classifiers, the number of fine-tuning epochs, learning rate, output dimensions of the sentence embeddings, pooling method, the inclusion of a normalization layer for embeddings, the holdout ratio of the training data, early stopping patience and threshold, loss function, fine-tuning batch size, and whether to return the pooled BERT embeddings directly or add feedforward layers, among others
(see Appendix~\ref{subsec:ablation} for results).

After exploring these hyperparameters, we obtained an accuracy of 88.83\%, which stands as our best result. This model was trained with McEMcC classifier, triplet-soft loss, the Sentence-BERT mean-pooled embeddings, neural network classifiers, 10 fine-tuning epochs, fine-tuning batch size of 16, no early-stopping and holdout ratio of 0.0, no normalization layer, and using the 768-dimensional pooled BERT embeddings directly. 

We compared our results with the results reported in the AEGIS paper \cite{ghosh2024}, reproduced in Table~\ref{Model Results Comparison}. We see that despite our final solution only containing 67M parameters in total, we are able to perform on par with significantly larger models (in excess of 7 billion parameters). 

\begin{table}[tb]
\centering
\resizebox{0.5\textwidth}{!}{
\begin{tabular}{lcc}
\hline
\textbf{}                     & \multicolumn{2}{c}{\textbf{AEGIS (on-policy)}} \\
\textbf{}                     & \textbf{AUPRC}     & \textbf{F1}      \\
\hline
LlamaGuardBase (Meta)     & 0.930             & 0.62             \\
NeMo43B(Nvidia)              & -                 & 0.83             \\
OpenAI Mod API       & 0.895             & 0.34             \\
Perspective API      & 0.860             & 0.24             \\
\hline
LlamaGuardDefensive (AEGIS) & 0.941             & 0.85             \\
LlamaGuardPermissive (AEGIS) & 0.941             & 0.76             \\
NeMo43B-Defensive (AEGIS)   & -                 & 0.89             \\
WildGuard (most recent) &  - & 0.89 \\
\hline
\textbf{Our Sentence-BERT model} & \textbf{0.946} & \textbf{0.89}
\\
\hline
\end{tabular}
}
\caption{Comparison of \citet{ghosh2024} and our model's results on the AEGIS test data. AUPRC = area under the precision-recall curve.}
\label{Model Results Comparison}
\end{table}

\begin{figure}[tb]
    \centering
    \includegraphics[width=0.5\textwidth]{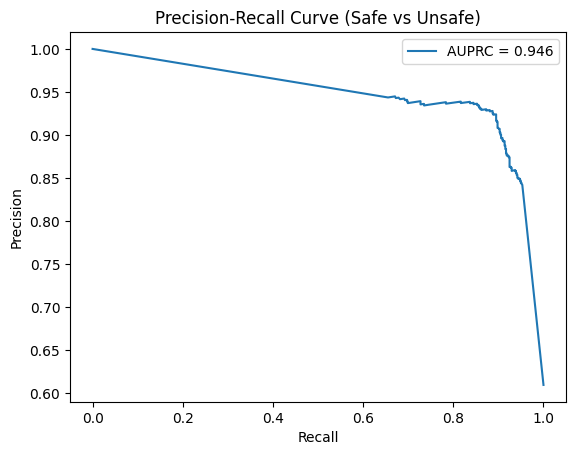} 
    \caption{Precision-recall curve for our best model}
    \label{fig:AUPRC Curve}
\end{figure}

The metric of area under the precision-recall curve (AUPRC) is illustrated in 
Figure~\ref{fig:AUPRC Curve}, which is a plot of the detection trade-off curve for our best-performing approach.  Note that for purposes of recall, the positive class is the unsafe category, which is what we aim to detect in most applications.

We tested our best model against the two models that were proposed by AEGIS and compared their latency. We observe that the fine-tuned LlamaGuard on a single GPU (g5.2xlarge AWS) instance has an inference latency time of over 140 seconds, while our best model has an inference latency time of about 0.05 seconds (see Appendix~\ref{subsec:inf_time_comp})

\section{Limitations and Future Work}

While the results are promising, there is is need for improvement in several directions. For starters, our investigation was for English only, and we have not yet explored fine-tuning Sentence-BERT on multilingual inputs. This restricts the effectiveness of our guardrail to English-speaking users, limiting its utility for global use. Second, our embedding model is constrained to text-based inputs, and does not accommodate other modalities, such as speech or video, which are increasingly common in interactions with LLMs. Finally, our current solution only provides generic unsafe input filtering and does not support few-shot topic-based filtering. As a result, application developers cannot define specific additional topics they wish to filter as unsafe, which limits the customization and flexibility of the guardrail. Addressing these limitations would significantly enhance the applicability and robustness of our system. Future work will explore ways in which to fine-tune an embedding model capable of both unsafe prompt filtering and few-shot topic filtering with minimal data.

Other directions for future investigation are suggested by observations of results already obtained.
We believe that our model's performance can be improved significantly, given that we only used a fraction of the AEGIS data.  After our data preprocessing, we retained 9,674 public annotated human-LLM interaction instances, compared to around 26,000 total instances in the corpus (see Appendix~\ref{subsec:category_dis}, Figure 4, for the full AEGIS category distribution).  When doing ablation studies varying the training dataset size, we found that there existed a monotonically increasing trend between the F1 score and the amount of training data used for embedding model (see Appendix~\ref{subsec:ablation}, or Table~\ref{Ablation Study Finetuning}). We can thus surmise that much better results could be obtained with substantially more data. This in turn suggests using techniques for data augmentation (e.g., paraphrasing) or machine-labeling (e.g., ensembling of powerful teacher models).

\section{Conclusion}

We have explored safety filters as an add-on to instruction-tuning heavy-duty LLM models, and introduced a new effective, lightweight guardrails approach. Our goal was to minimize the number of model parameters and reduce inference latency while retaining performance on the task of detecting unsafe LLM prompts. We have demonstrated a solution that involves fine-tuning a BERT-based model, using Sentence-BERT to learn embeddings  representing the safe/unsafe distinctions.  The learned embeddings are fed to a simple vector classifier for binary or multi-class categorization. We found that retaining distinct unsafe categories for both embedding training and embedding classification yielded the best overall results.
The final results are comparable to popular LLM approaches based on models many orders of magnitude larger than ours, making this approach suitable for low-cost integration into varied LLM applications.


\bibliography{custom.bib}

\begin{thebibliography}{15}
\providecommand{\natexlab}[1]{#1}

\bibitem[{Brin and Page(1998)}]{brin1997}
Sergey Brin and Lawrence Page. 1998.
\newblock The anatomy of a large-scale hypertextual web search engine.
\newblock \emph{Computer networks and ISDN systems}, 30(1-7):107--117.

\bibitem[{Broder(2002)}]{broder2002}
Andrei Broder. 2002.
\newblock A taxonomy of web search.
\newblock \emph{ACM SIGIR Forum}, 36(2):3--10.

\bibitem[{Cutting et~al.(1993)Cutting, Karger, and Pedersen}]{cutting1993}
Douglass~R Cutting, David~R Karger, and Jan~O Pedersen. 1993.
\newblock Constant interaction-time scatter/gather browsing of very large document collections.
\newblock In \emph{Proc.\ 16th Annual Intl.\ ACM SIGIR Conference on Research and Development in Information Retrieval}, pages 126--134.

\bibitem[{Ghosh et~al.(2024)Ghosh, Varshney, Galinkin, and Parisien}]{ghosh2024}
Shaona Ghosh, Prasoon Varshney, Erick Galinkin, and Christopher Parisien. 2024.
\newblock {AEGIS}: Online adaptive {AI} content safety moderation with ensemble of {LLM} experts.
\newblock arXiv preprint arXiv:2404.05993.

\bibitem[{Han et~al.(2024)Han, Rao, Ettinger, Jiang, Lin, Lambert, Choi, and Dziri}]{han2024}
Seungju Han, Kavel Rao, Allyson Ettinger, Liwei Jiang, Bill~Yuchen Lin, Nathan Lambert, Yejin Choi, and Nouha Dziri. 2024.
\newblock {WildGuard}: Open one-stop moderation tools for safety risks, jailbreaks, and refusals of {LLMs}.
\newblock arXiv preprint arXiv:2406.18495.

\bibitem[{{Huggingface}(2024)}]{huggingface}
{Huggingface}. 2024.
\newblock Pretrained models.
\newblock https://huggingface.co/transformers/v2.9.1/pretrained\_models.html.

\bibitem[{HuggingFace(2024)}]{sentence-transformers}
HuggingFace. 2024.
\newblock Sentence transformers.
\newblock https://huggingface.co/sentence-transformers.

\bibitem[{Inan et~al.(2023)Inan, Upasani, Chi, Rungta, Iyer, Mao, Tontchev, Hu, Fuller, Testuggine et~al.}]{inan2024}
Hakan Inan, Kartikeya Upasani, Jianfeng Chi, Rashi Rungta, Krithika Iyer, Yuning Mao, Michael Tontchev, Qing Hu, Brian Fuller, Davide Testuggine, et~al. 2023.
\newblock {LlamaGuard}: {LLM}-based input-output safeguard for human-{AI} conversations.

\bibitem[{Kenton and Toutanova(2019)}]{kenton2019bert}
Jacob Devlin Ming-Wei~Chang Kenton and Lee~Kristina Toutanova. 2019.
\newblock {BERT}: Pre-training of deep bidirectional transformers for language understanding.
\newblock In \emph{Proceedings of NAACL-HLT}, volume~1, page~2.

\bibitem[{Lees et~al.(2022)Lees, Tran, Tay, Sorensen, Gupta, Metzler, and Vasserman}]{lees2022}
Alyssa Lees, Vinh~Q Tran, Yi~Tay, Jeffrey Sorensen, Jai Gupta, Donald Metzler, and Lucy Vasserman. 2022.
\newblock A new generation of {P}erspective {API}: Efficient multilingual character-level transformers.
\newblock In \emph{Proceedings of the 28th ACM SIGKDD conference on knowledge discovery and data mining}, pages 3197--3207.

\bibitem[{OpenAI(2024)}]{openaiAPI}
OpenAI. 2024.
\newblock {OpenAI API}.
\newblock https://platform.openai.com/docs/guides/moderation/overview.

\bibitem[{Parisien(2024)}]{aegis-personal}
Christopher Parisien. 2024.
\newblock Personal communication.

\bibitem[{Rebedea et~al.(2023)Rebedea, Dinu, Sreedhar, Parisien, and Cohen}]{rebedea2023}
Traian Rebedea, Razvan Dinu, Makesh Sreedhar, Christopher Parisien, and Jonathan Cohen. 2023.
\newblock {NeMo Guardrails}: A toolkit for controllable and safe {LLM} applications with programmable rails.

\bibitem[{Reimers and Gurevych(2019)}]{reimers2019}
Nils Reimers and Iryna Gurevych. 2019.
\newblock {Sentence-BERT}: Sentence embeddings using {Siamese} {BERT}-networks.
\newblock In \emph{Proc.\ Conf.\ Empirical Methods in Natural Language Processing and 9th Intl.\ Joint Conf.\ on Natural Language Processing (EMNLP-IJCNLP)}, pages 3982--3992, Hong Kong.

\bibitem[{Salton(1989)}]{salton1989}
Gerald Salton. 1989.
\newblock \emph{Automatic text processing: The transformation, analysis, and retrieval of information by computer}.
\newblock Addison-Wesley.

\end{thebibliography}


\section{Appendix}

\subsection{Inference time comparison with LlamaGuard}
\label{subsec:inf_time_comp}

\begin{table}[htbp]
\centering
\resizebox{0.5\textwidth}{!}{ 
\begin{tabular}{lccc}
\toprule
\textbf{Model} & \textbf{Iter} & \textbf{Inf time} & \textbf{StDev of inf time} \\
                &           &  (sec)            & (sec) \\
\midrule
\multirow{6}{*}{BERT-based} & 1 & \large 0.0522 & \large 0.1386 \\
                            & 2 & \large 0.0551 & \large 0.1473 \\
                            & 3 & \large 0.0493 & \large 0.1296 \\
                            & 4 & \large 0.1037 & \large 0.2779 \\
                            & 5 & \large 0.0585 & \large 0.1572 \\
                            & 6 & \large 0.0573 & \large 0.1532 \\
\midrule
\multirow{2}{*}{LG Permissive}      & 1 & \large 163.6732 & \large 5.9813 \\
                            & 2 & \large 162.4430 & \large 5.9557 \\
\midrule
\multirow{2}{*}{LG Defensive}      & 1 & \large 164.3122 & \large 5.6623 \\
                            & 2 & \large 162.3445 & \large 5.2810 \\
\bottomrule
\end{tabular}
}
\caption{Inference times for our BERT-based and the LlamaGuard models}
\label{tab:inference_times}
\end{table}

\subsection{Hyperparameters used for evaluation}
\label{subsec:hyperparameter_lis}

\begin{center}
\begin{tabular}{|l|c|}
\hline
\textbf{Parameter}            & \textbf{Value}         \\ \hline
Holdout ratio                 & 0.0                   \\ \hline
Add normalization             & True                  \\ \hline
Classifier                    & SVM                   \\ \hline
Add feedforward               & True                  \\ \hline
Fine-tuning batch size         & 16                    \\ \hline
Random seed                   & 21                    \\ \hline
Fine-tuning epochs             & 10                    \\ \hline
Embedding dimension           & 768                   \\ \hline
Loss function                 & Triplet-soft          \\ \hline
Pooling method                & Mean                  \\ \hline
\end{tabular}
\end{center}

\subsection{AEGISSafetyDataset category distribution}
\label{subsec:category_dis}

 \begin{center} 
 \vspace{-1em}
    \includegraphics[width=0.5\textwidth]{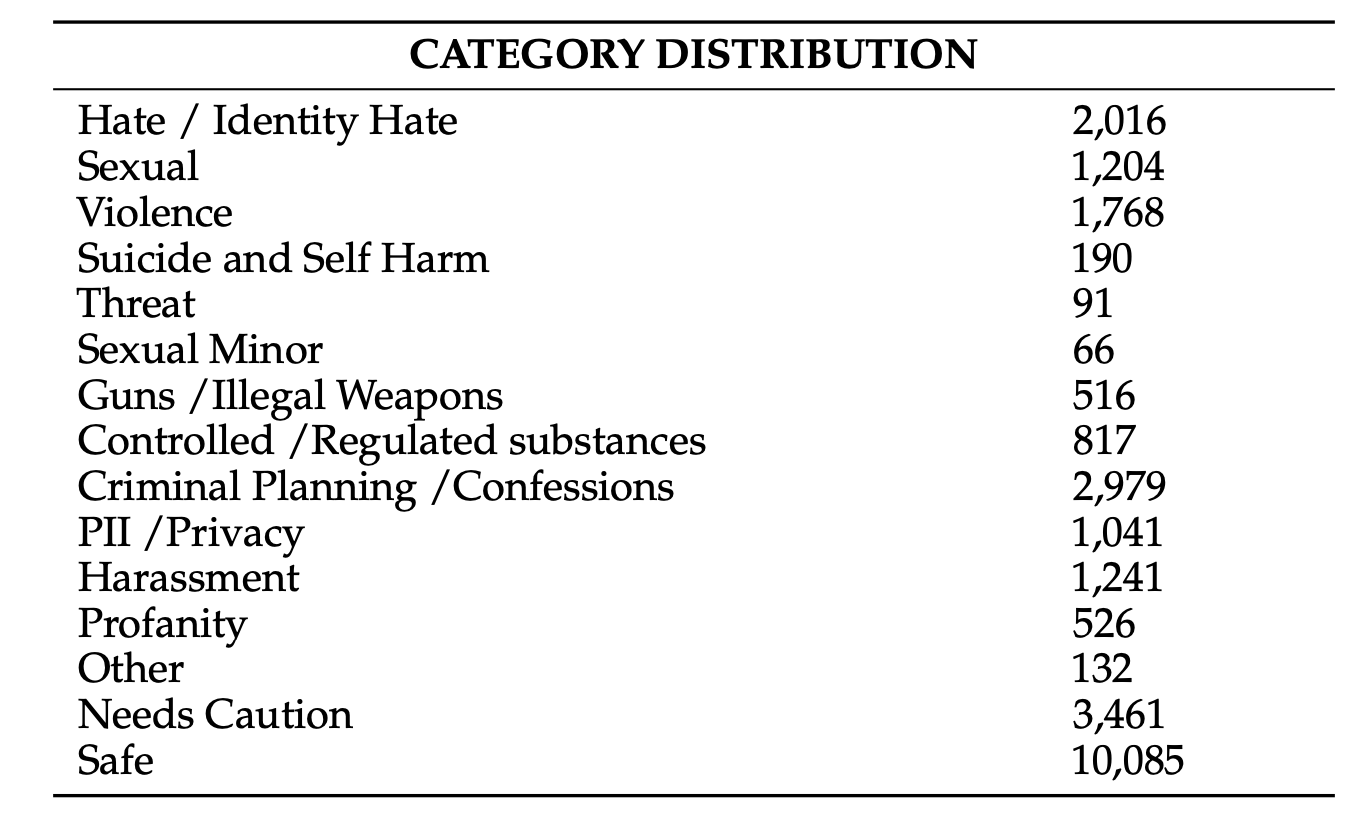} 
 \end{center}

\subsection{Ablation tests}
\label{subsec:ablation}
We first list the default hyperparameter values used in our ablation tests.
\begin{table}[h]
    \centering
    \begin{tabular}{|l|c|}
        \hline
        \textbf{Parameter}      & \textbf{Value} \\ \hline
        Approach                & 1             \\ \hline
        Holdout ratio           & 0.0           \\ \hline
        Add normalization       & True          \\ \hline
        Classifier              & SVM           \\ \hline
        Add feedforward         & True          \\ \hline
        Fine-tuning batch size   & 16            \\ \hline
        Fine-tuning epochs       & 10            \\ \hline
        Embedding dimension     & 512           \\ \hline
        Loss function           & Triplet-soft  \\ \hline
        Pooling method          & Mean          \\ \hline
    \end{tabular}
\end{table}

Next, we report ablations for several of the parameters, as indicated in the table captions below.
Each reported result is an average over runs with four different seed values, except the ablation study for fine-tuning ratio (Table~\ref{Ablation Study Finetuning}), where we average over six seed values. 

\begin{table}[htbp]
\centering
\resizebox{0.5\textwidth}{!}{%
\begin{tabular}{|c|c|c|c|c|}
\hline
\textbf{Fine-tuning ratio} & \textbf{Accuracy} & \textbf{F1} & \textbf{UAP} \\ \hline
0.2                       & 0.839280          & 0.815721    & 0.822517    \\ \hline
0.4                       & 0.847798          & 0.830100    & 0.834376    \\ \hline
0.6                       & 0.858406          & 0.844764    & 0.847707    \\ \hline
0.8                       & 0.870781          & 0.861326    & 0.862864    \\ \hline
1.0                       & 0.877853          & 0.872465    & 0.873006    \\ \hline
\end{tabular}%
}
\caption{Ablation: fine-tuning ratio, i.e., percentage of available data used for fine-tuning of embedding model}
\label{Ablation Study Finetuning}
\end{table}

\begin{table}[htbp]
\centering
\resizebox{0.5\textwidth}{!}{%
\begin{tabular}{|l|c|c|c|c|}
\hline
\textbf{}        & \textbf{Triplet soft} & \textbf{Triplet hard} & \textbf{Triplet all} & \textbf{Contrastive} \\ 
\hline
\textbf{Accuracy (\%)}    & 87.56  & 87.75    & 88.24              & 87.46              \\ \hline
\textbf{F1 score (\%)}    & 86.63              & 86.65             & 87.51              & 87.08              \\ \hline
\textbf{UAP (\%)}         & 86.77              & 86.84             & 87.60              & 87.11              \\ \hline
\end{tabular}%
}
\caption{Ablation: different loss functions}
\label{Ablation Study Loss Functions}
\end{table}

\begin{table}[htbp]
\centering
\resizebox{0.5\textwidth}{!}{%
\begin{tabular}{|l|c|c|c|c|}
\hline
\textbf{Embedding dimension} & \textbf{256}    & \textbf{512}    & \textbf{1024}   & \textbf{1536}   \\ \hline
\textbf{Accuracy (\%)}        & 87.95          & 87.56          & 87.46          & 88.33          \\ \hline
\textbf{F1 score (\%)}        & 87.23          & 86.63          & 86.43          & 87.34          \\ \hline
\textbf{UAP (\%)}             & 87.32          & 86.77          & 86.60          & 87.49          \\ \hline
\end{tabular}%
}
\caption{Ablation: dimension of embedding model}
\end{table}

\begin{table}[htbp]
\centering
\resizebox{0.5\textwidth}{!}{%
\begin{tabular}{|l|c|c|c|}
\hline
\textbf{Pooling strategy} & \textbf{MEAN}    & \textbf{MAX}     & \textbf{CLS}     \\ \hline
\textbf{Accuracy (\%)}    & 87.56            & 87.97           & 86.89           \\ \hline
\textbf{F1 score (\%)}    & 86.63            & 87.75           & 86.00           \\ \hline
\textbf{UAP (\%)}         & 86.77            & 87.76           & 86.13           \\ \hline
\end{tabular}%
}
\caption{Ablation: Sentence-BERT pooling strategy }
\end{table}

\begin{table}[htbp]
\centering
\resizebox{0.5\textwidth}{!}{%
\begin{tabular}{|l|c|c|c|c|}
\hline
\textbf{Fine-tuning epochs} & \textbf{3}       & \textbf{5}       & \textbf{10}      & \textbf{20}      \\ \hline
\textbf{Accuracy (\%)}     & 86.40            & 87.46            & 87.56            & 87.46            \\ \hline
\textbf{F1 score (\%)}     & 85.12            & 86.56            & 86.63            & 86.62            \\ \hline
\textbf{UAP (\%)}          & 85.37            & 86.69            & 86.77            & 86.74            \\ \hline
\end{tabular}%
}
\caption{Ablation: fine-tuning epochs}
\end{table}

\begin{table}[htbp]
\centering
\resizebox{0.5\textwidth}{!}{%
\begin{tabular}{|l|c|c|}
\hline
\textbf{Normalization}         & \textbf{No Normalization} & \textbf{Normalization} \\ \hline
\textbf{Accuracy (\%)}         & 87.56                     & 87.46                  \\ \hline
\textbf{F1 score (\%)}         & 86.63                     & 86.68                  \\ \hline
\textbf{UAP (\%)}              & 86.77                     & 86.79                  \\ \hline
\end{tabular}%
}
\caption{Ablation: normalization of vector embeddings}
\end{table}

\FloatBarrier

\end{document}